\documentclass[conference]{IEEEtran}
\IEEEoverridecommandlockouts
\usepackage{cite}
\usepackage{amsmath,amssymb,amsfonts}
\usepackage{algorithmic}
\usepackage{graphicx}
\usepackage{textcomp}
\usepackage{xcolor}
\usepackage{url}
\usepackage{subcaption}
\usepackage{comment}

\begin{document}

\title{Profiling AI Models: Towards Efficient Computation Offloading in Heterogeneous Edge AI Systems}

\author{%
\IEEEauthorblockN{Juan Marcelo Parra-Ullauri}
\IEEEauthorblockA{\textit{Smart Internet Lab} \\
\textit{University of Bristol}\\
Bristol, UK \\
jm.parraullauri@bristol.ac.uk}\and
 \IEEEauthorblockN{Oscar Dilley}
\IEEEauthorblockA{\textit{Smart Internet Lab} \\
\textit{University of Bristol}\\
Bristol, UK \\
oscar.dilley@bristol.ac.uk}\and
 \IEEEauthorblockN{Hari Madhukumar}
\IEEEauthorblockA{\textit{Smart Internet Lab} \\
\textit{University of Bristol}\\
Bristol, UK \\
h.madhukumar@bristol.ac.uk}\and
 \IEEEauthorblockN{Dimitra Simeonidou}
\IEEEauthorblockA{\textit{Smart Internet Lab} \\
\textit{University of Bristol}\\
Bristol, UK \\
dimitra.simeonidou@bristol.ac.uk}

\thanks{This work is a contribution by Project REASON, a UK Government funded project under the Future Open Networks Research Challenge (FONRC) sponsored by the Department of Science Innovation and Technology (DSIT).}
	}
\maketitle
\vspace{-0.2cm}
\begin{abstract}
The rapid growth of end-user AI applications, such as computer vision and generative AI, has led to immense data and processing demands often exceeding user devices' capabilities. Edge AI addresses this by offloading computation to the network edge, crucial for future services in 6G networks. However, it faces challenges such as limited resources during simultaneous offloads and the unrealistic assumption of homogeneous system architecture. To address these, we propose a research roadmap focused on profiling AI models, capturing data about model types, hyperparameters, and underlying hardware to predict resource utilisation and task completion time. Initial experiments with over 3,000 runs show promise in optimising resource allocation and enhancing Edge AI performance.

\end{abstract}

\vspace{-0.2cm}
\section{Introduction}
The rapid growth of end-user AI applications, such as real-time image recognition and generative AI, has led to high data and processing demands that often exceed device capabilities. Edge AI addresses these challenges by offloading computation to the network's edge, where hardware-accelerated AI processing can occur~\cite{laskaridis2024FutureEdgeAI}. This approach is integral to AI and RAN, a key component of future 6G networks as outlined by the AI-RAN Alliance\footnote{\url{https://ai-ran.org/working-groups/}}. In 6G, AI integration across edge-RAN and extreme-edge devices will support efficient data distribution and distributed AI techniques, enhancing privacy and reducing latency for applications like the Metaverse and remote surgery. Despite these benefits, Edge AI faces challenges. Limited resource availability at the edge can hinder performance during simultaneous offloads. Additionally, the assumption of homogeneous system architecture in the existing literature is unrealistic, as edge devices vary widely in processor speeds and architectures (e.g., 1.5GHz vs 3.5GHz, or X86 vs ARM), impacting task processing and resource utilisation.

To address these challenges, we propose a research roadmap focused on profiling AI models by analysing their execution dynamics across various bare-metal systems. Our goal is to understand how AI model types (e.g., MLP, CNN), hyperparameters (e.g., learning rate, optimiser), hardware (e.g., architecture, FLOPS), and dataset characteristics (e.g., size, batch size) affect model accuracy, resource use, and task completion time. This \emph{Profiling AI Models} process allows us to predict resource needs and task completion times, enabling efficient scheduling across edge nodes. Our initial experiments, involving over 3,000 runs with varied configurations, showcase the effectiveness of our approach. Using AI techniques like XGBoost, we achieved a normalised RMSE of 0.001, a significant improvement over MLPs with over 4 million parameters.

\begin{figure}
    \centering
    \includegraphics[width=\linewidth]{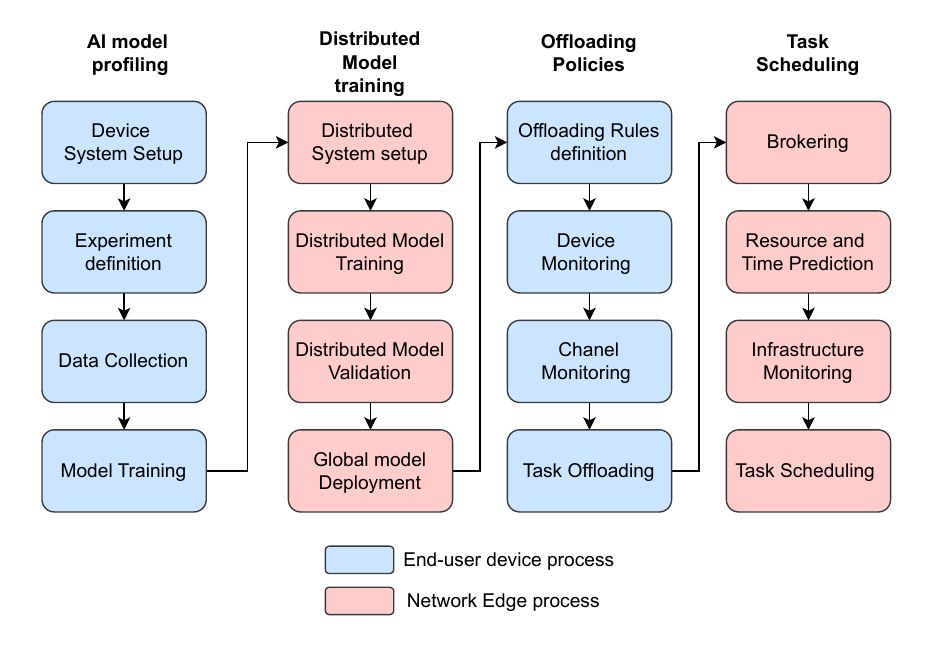}
    \caption{Research roadmap for profiling based computation offloading}
    \label{fig:roadmap}
\end{figure}
\vspace{-0.1cm}
\section{Roadmap Towards Efficient Computation Offloading in Heterogeneous Edge AI Systems}

\subsection{Local AI model Profiling}
This stage focuses on analysing how device dynamics and system characteristics influence the performance of AI models across varied hardware setups and environments. The goal is to uncover relationships among factors such as AI model types (MLP, CNN), hyperparameters (learning rate, optimiser), hardware specifications (architecture, FLOPS), and dataset characteristics (size, batch size), and their effects on model accuracy, resource utilisation, and task completion time. Profiling these elements helps predict system behaviour under different configurations and workloads, which is crucial for optimising resource usage and ensuring service quality~\cite{zhangb2024federated}.

The process begins with setting up device systems and selecting appropriate probes and methods for characterisation. It involves choosing suitable systems for stress testing and defining experimental parameters, including model types, hyperparameters, datasets, processors and architectures.    Following this, the profiling process begins, leading to data collection to generate a dataset that captures the relationships between the various factors. Once this dataset is created, different types of regression models are trained to benchmark and determine which model best predicts resource usage and task completion time for specific AI tasks. This iterative process can be conducted across any underlying systems, models, and datasets. The goal is to develop diverse profiling models that accurately reflect system performance and resource demands. We will provide a more detailed description of this stage and present our initial results in Section~\ref{use_case}.

\subsection{Distributed Model training}
In the previous stage, individual models are created to characterise data dependencies specific to their hardware. The next challenge is to generalise across diverse systems. This complexity arises from distributed data and privacy concerns, as user data may be sensitive~\cite{pang2022towards}. To address these issues, we propose using Federated Learning (FL) and differential privacy, based on our previous work~\cite{parra2024kubeflower}. Our framework utilises cloud-native technologies and principles of isolation by design and differential privacy to ensure secure FL.

This stage involves setting up the distributed system, including defining the FL pipeline, configuring the server, selecting aggregation methods, and establishing communication schemes between clients and the server. Distributed model training follows system configuration. Validation can be conducted through federated validation, where each client uses a holdout test dataset, or centralised validation, where the server tests the global model with an unseen dataset. Once validated, the model is packaged and deployed as a global profiling model to predict resources required for offloaded tasks.

\subsection{Offloading Policies}
Computation offloading in Edge AI has been extensively studied, but there are still areas requiring further research, such as optimising timing and location for offloading~\cite{AKHLAQI2023103568}. A prominent method is split computing, where edge devices offload parts of neural network computations to nearby servers, reducing latency. Deep Reinforcement Learning (DRL) algorithms typically govern which neural network layers to offload. After making an offloading decision, control shifts to the edge server, where the component is executed as scheduled.

The offloading process involves several steps: defining offloading rules, monitoring the device’s state and capabilities, assessing link conditions, and executing the task offloading. These steps ensure that the offloading is efficient and adapts to the dynamic conditions of edge environments.

\subsection{Task Scheduling}

Task scheduling in Edge AI involves not only selecting the most suitable edge node for process execution but also determining the optimal sequence of tasks. Advanced scheduling algorithms aim to predict future system states and pre-allocate resources to enhance long-term performance. Accurate task duration estimates are crucial, especially for AI workloads. Profilers play a key role by predicting Pareto-optimal resource and time combinations for tasks, ensuring compliance with latency and Quality of Service (QoS) requirements. This allows schedulers to make informed decisions, thus improving overall system performance.

The scheduling process includes several stages: task brokering, which involves managing and prioritising user-offloaded tasks; resource and time prediction, using global profiling models to estimate computational needs and task duration; and infrastructure monitoring, assessing the availability of network and edge nodes. Based on this data, optimisation techniques are used for task scheduling, placement, and execution. The scheduling problem may be modelled as a Markov Decision Process (MDP) or a Partially Observable (PO)-MDP, depending on the completeness of state information from all nodes, to address uncertainties in dynamic edge environments.

\begin{table}[h]
\centering
\begin{tabular}{|l|l|}
\hline
\textbf{Type of Variable}     & \textbf{Configurations} \\ \hline
\textbf{CNN Types}            & 
\begin{tabular}[c]{@{}l@{}}
- \{out\_channels: 32, kernel\_size: 5, pool: True\} \\
- \{out\_channels: 32, kernel\_size: 5, pool: True\}, \\
   \{out\_channels: 64, kernel\_size: 3, pool: True\} \\
- \{out\_channels: 32, kernel\_size: 5, pool: True\}, \\
   \{out\_channels: 64, kernel\_size: 3, pool: True\}, \\
   \{out\_channels: 128, kernel\_size: 3, pool: True\} \\
\end{tabular} \\ \hline

\textbf{MLP Types}            & 
\begin{tabular}[c]{@{}l@{}}
- [100, 50] \\
- [150, 100, 50] \\
- [200, 150, 100, 50] \\
\end{tabular} \\ \hline

\textbf{Epochs}               & 5, 10, 15, 20 \\ \hline
\textbf{Optimisers}           & Adam, SGD, RMSprop, Adagrad \\ \hline
\textbf{Learning Rates}       & 0.01, 0.05, 0.001, 0.005, 0.0001, 0.0005 \\ \hline
\textbf{Batch Sizes}          & 16, 32, 64, 128 \\ \hline
\end{tabular}
\vspace{0.1cm}
\caption{Configurations for CNN, MLP, and Hyperparameters for Profiling dataset generation}
\label{tab:configurations}
\end{table}
\vspace{-0.2cm}
\section{Example Use case}
\label{use_case}

\subsection{Scenario description}
Throughout this document, we emphasise that Edge AI in 6G networks involves a variety of heterogeneous devices, which complicates the scheduling of offloaded AI tasks. This section details the initial experiments from the first stage of our research roadmap, focusing on on-device AI model profiling. The goal of the \emph{AI profiling} phase is to determine viable models for profiling AI tasks based on hardware characteristics and model hyperparameters. We collected dataset by training models and profiling them on a Dell XPS 15 with an Intel Core i5 and NVIDIA GeForce GTX 1650 GPU, as shown in Table \ref{tab:configurations}. We compared MLP regression with XGBoost, which excels in handling tabular data. Initial results are promising, and the next phase involves comparing the performance of these models trained in a distributed setting.

\subsection{Experimental results}
\begin{figure}[h!]
\centering
\begin{subfigure}[b]{0.75\linewidth}
         \centering
         \includegraphics[width=\linewidth]{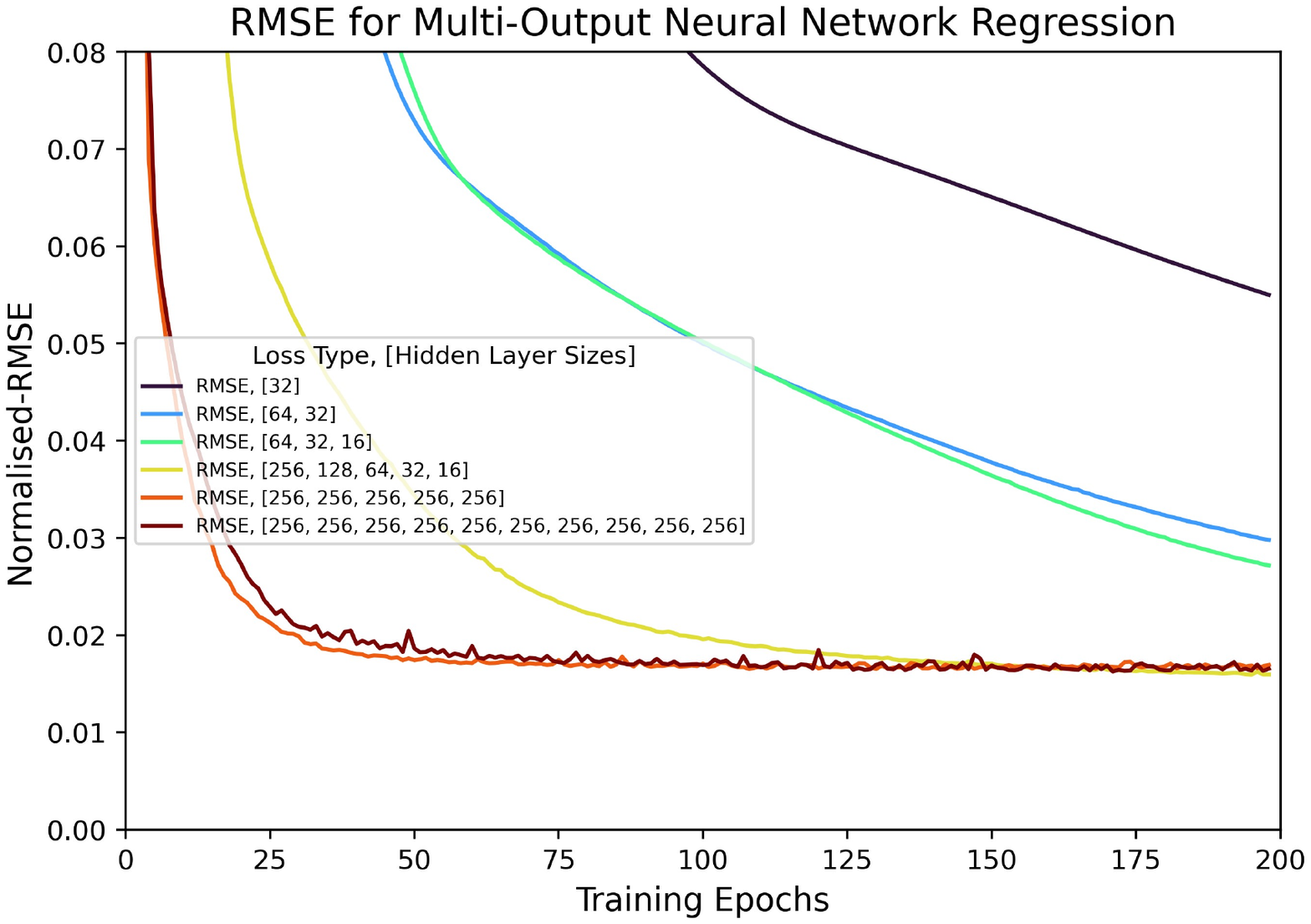}
         \caption{\centering MLP regression models. Individual models for each target are stacked. Models have between 3,143 and 4,169,991 parameter}
         \label{fig:nn}
\end{subfigure}
\begin{subfigure}[b]{0.75\linewidth}
         \centering
         \includegraphics[width=\linewidth]{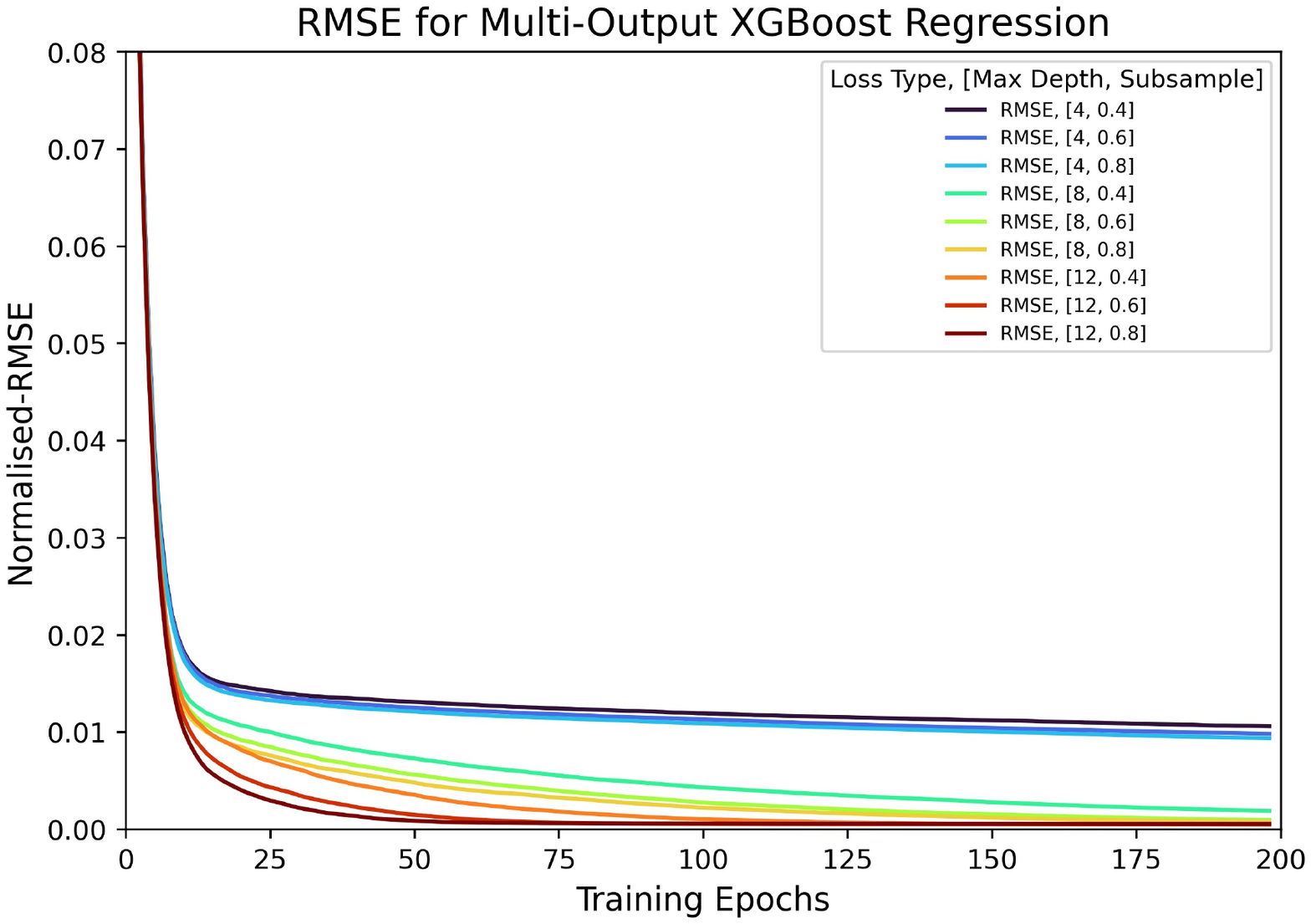}
         \caption{\centering XGBoost regression models. An individual boosted tree ensemble is used for each target.}
         \label{fig:xgboost}
\end{subfigure}
\caption{Comparing error performance of different models for AI profiling.}
\label{fig:performance}
\end{figure}

\begin{figure}
    \centering
    \includegraphics[width=0.68\linewidth]{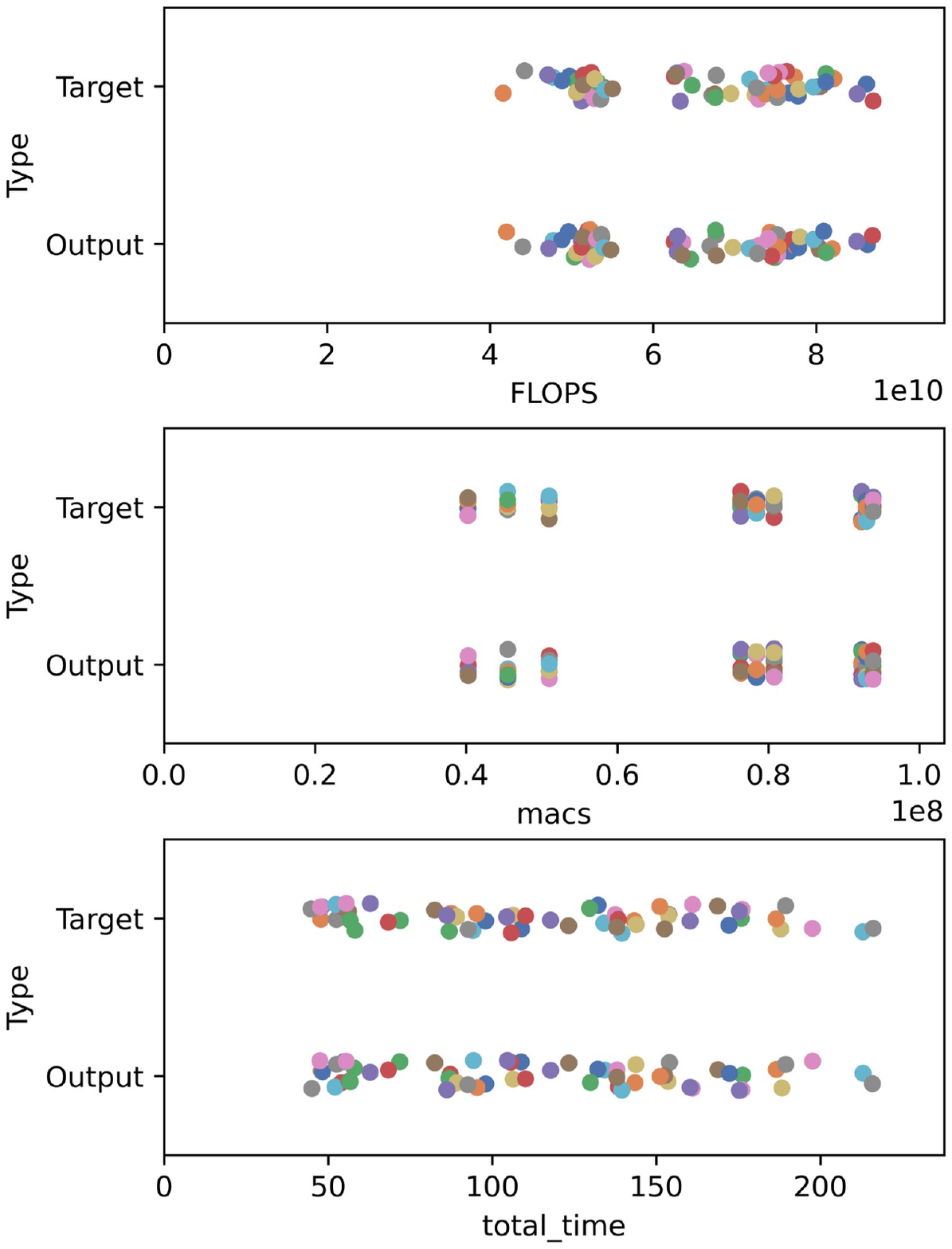}
    \caption{Demonstrating the performance of the XGBoost (max depth=12, subsample=0.8) model in predicting FLOPS, macs and total time.}
    \label{fig:swarmplot}
\end{figure}

The results in Figure \ref{fig:performance} compare the RMSE performance of the NN and XGBoost-based models for predicting the model profile. Figure \ref{fig:nn} demonstrates that increasing the number of parameters in the model (by increasing the depth and width of the hidden layer) improves the convergence time. Further, it also improves the model's accuracy, up to an irreducible error floor experienced at a normalised RMSE just below 0.02. The performance of the large MLPs is predicted to improve further with more dataset entries but there are concerns about the generality of the model and overfitting with large numbers of parameters.

The max-depth of tree ensembles and sub-sample rate are varied to compare XGBoost models. Optimal tree-based models outperform the largest MLP (4,169,991 parameters) by an order of magnitude. These results, shown in Figure \ref{fig:xgboost}, suggest that max-depth and sub-sample rate are proportionate to model accuracy. The effect of max-depth is more significant but offers diminishing returns beyond a certain point. Figure \ref{fig:swarmplot} shows the denormalised predictions of the highest performing XGBoost model compared to the target, highlighting its accuracy. These findings suggest decision tree-based models may be more effective for AI profiling. However, due to the unpredictable effects of distributed learning on model performance, a range of models will be considered in the next research stage.

\section{Conclusion and Future Work}

We present a comprehensive roadmap and a robust framework for profiling AI models in Edge AI systems. By capturing and analysing data on model types, hyperparameters, and hardware architecture, we can accurately predict resource utilisation and task completion times across various edge nodes. Our preliminary experiments, involving over 3,000 runs, validate the effectiveness and efficiency of this approach, leading to substantial improvements towards resource allocation and task scheduling. The advancements observed, including a notable reduction in error metrics with XGBoost, highlight the potential for optimising Edge AI performance in future 6G networks. We will focus on the subsequent stages of our research roadmap, which will build upon these findings to further enhance Edge AI Systems capabilities.

\bibliographystyle{IEEEtran} 
\scriptsize{\bibliography{IEEEabrv,citations}}
\end{document}